\documentclass[conference]{IEEEtran}
\IEEEoverridecommandlockouts
\usepackage{cite}
\usepackage{amsmath,amssymb,amsfonts}
\usepackage{algorithmic}
\usepackage{graphicx}
\usepackage{textcomp}
\usepackage{xcolor}
\usepackage{cite}
\newtheorem{definition}{Definition}
\usepackage[scientific-notation=true]{siunitx}
\usepackage[numbers]{natbib} 
\usepackage[ruled,vlined,linesnumbered]{algorithm2e}
\def\BibTeX{{\rm B\kern-.05em{\sc i\kern-.025em b}\kern-.08em
    T\kern-.1667em\lower.7ex\hbox{E}\kern-.125emX}}

\usepackage{setspace}

\begin{document}

\title{Encoder-Decoder Generative Adversarial Nets for Suffix Generation and Remaining Time Prediction of Business Process Models
}

\author{\IEEEauthorblockN{ Farbod Taymouri}
\IEEEauthorblockA{
\textit{The University of Melbourne}\\
Melbourne, Australia \\
0000-0002-3150-336X}
\and

\IEEEauthorblockN{ Marcello La Rosa}
\IEEEauthorblockA{
\textit{The University of Melbourne}\\
Melbourne, Australia \\
0000-0001-9568-4035}


}

\maketitle
\thispagestyle{plain}
\pagestyle{plain}

\begin{abstract}
Predictive process monitoring aims to predict future characteristics of an ongoing process case, such as case outcome or remaining time till completion. Several deep learning models have been proposed to address suffix generation and remaining time prediction for ongoing process cases. Though they generally increase the prediction accuracy compared to traditional machine learning models, they still suffer from critical issues. For example, suffixes are generated by training a model on iteratively predicting the next activity. As such, prediction errors are propagated from one prediction step to the next, resulting in poor reliability, i.e., the ground truth and the generated suffixes may easily become dissimilar. Also, conventional training of neural networks via maximum likelihood estimation is prone to overfitting and prevents the model from generating sequences of variable length and with different activity labels. This is an unrealistic simplification as business process cases are often of variable length in reality. 
To address these shortcomings, this paper proposes an encoder-decoder architecture grounded on Generative Adversarial Networks (GANs), that generates a sequence of activities and their timestamps in an end-to-end way. GANs work well with differentiable data such as images. However, a suffix is a sequence of categorical items. To this end, we use the Gumbel-Softmax distribution to get a differentiable continuous approximation.
The training works by putting one neural network against the other in a two-player game (hence the ``adversarial'' nature), which leads to generating suffixes close to the ground truth. From the experimental evaluation it emerges that the approach is superior to the baselines in terms of the accuracy of the predicted suffixes and corresponding remaining times, despite using a naive feature encoding and only engineering features based on control flow and events completion time.
\end{abstract}


\section{Introduction}
\label{sec: introduction}

Predictive business process monitoring is an area of process mining that is concerned with predicting future characteristics of an ongoing process case. Different machine learning techniques, and more recently deep learning methods, have been employed to deal with different prediction problems,
such as outcome prediction \cite{TeinemaaDLM18}, remaining time prediction \cite{Tax17}, next event prediction \cite{Taymouri2020PredictiveBP, EVERMANN2017129,Tax17,Lin2019MMPredAD,Camargo2019LearningAL,Pasquadibisceglie2019UsingCN}, or suffix prediction (i.e.\ predicting the most likely continuation of an ongoing case) \cite{Tax17,Lin2019MMPredAD,Camargo2019LearningAL}.
In this paper, we are specifically interested in the latter problem: given an ongoing process case, called the \textit{prefix}, and an event log of completed cases for the same business process, we want to predict the most likely continuation for that prefix, by determining the sequence of event labels (process activities), called the \textit{suffix}, and the corresponding remaining time until the case finishes.
This problem has been addressed in \cite{EVERMANN2017129,Tax17,Lin2019MMPredAD,Camargo2019LearningAL} using Recurrent Neural Networks (RNNs) with Long-Short-Term Memory (LSTM). 

These approaches generally strike higher levels of prediction accuracy compared to the use of traditional machine learning models. Yet, they suffer from some issues. For example, a neural network architecture is trained to maximize its accuracy for the next activity prediction. A suffix is generated by simply predicting the next activity iteratively. This approach propagates the error from one step to the next steps resulting in poor suffix quality. Besides, all the proposed methods are \emph{discriminative}, i.e., they generate a suffix by iteratively classifying the next activity, whereas the suffix generation task requires a \emph{generative} model that approximates the underlying probability distribution which generates ground truth sequences. As such, these approaches tend to generate similar sequences, both in terms of length (fixed) as well as in terms of activity labels (limited set). These are unrealistic assumptions in the realm of business processes, where cases are of varied length and characterized by different activities.


Motivated by recent developments in the area of Generative Adversarial Nets (GANs) \cite{GANNIPS2014_5423}, this paper proposes a novel encoder-decoder generative adversarial framework to address the problem of suffix generation and remaining time prediction. The encoder-decoder architecture allows one to learn a mapping from prefixes to suffixes with variable lengths in an end-to-end way. Thereby, the model is trained on the whole ground truth suffixes directly. As such, the model can learn the relationships and orders between activities in prefixes and suffixes simultaneously. In our framework,
we call the encoder-decoder architecture as a generator, and put it against another neural network, called discriminator, in an adversarial \emph{minmax game} such that each network's goal is to maximize its own outcome at the cost of minimizing the opponent's outcome. One network generates the suffix and the remaining time, while the other network determines how realistic they are. Training continuous until the predictions are almost indistinguishable from the ground truth. During training, one player learns how to generate sequences of events close to the training sequences iteratively. 


To the best of our knowledge, this is the first work that adapts GANs for suffix generation and remaining time prediction for predictive process monitoring. This approach comes with several advantages. First, we show that our framework systematically outperforms the baselines in terms of accuracy, despite only using the order and completion time of events as features, as opposed to some of the baselines, which also use other available attributes in the log, such as the resource associated with each process activity. Second, our framework provides the $k$ most probable suffixes for a prefix that can be used for further analysis such as process forecasting \cite{ProcessForecastingquteprints131983}. 


We instantiated our framework using an LSTM architecture, and a naive one-hot encoding of the event labels in the log. Using this implementation, we evaluated the accuracy of our approach experimentally against three baselines targeted at the same prediction problem, using real-life event logs. 

The rest of this paper is organized as follows. The background and related work are provided in Sec. \ref{sec: back and related work}. The approach is described in Sec. \ref{sec:proposed approach} while the evaluation is discussed in Sec.\ref{sec: evaluation}. Finally, Sec. \ref{sec: conclusion} concludes the paper and discusses opportunities for future work.

\section{Background and Related work}
\label{sec: back and related work}
\noindent In this section we provide the required background knowledge. Next, we discuss related work in predictive process monitoring, with a focus on predicting the future of a running case using deep learning.

\subsection{Machine learning and Deep Learning}
\label{subsec: ml}
\noindent The goal of \emph{machine learning} is to develop methods that can automatically detect patterns in data to perform decision making under uncertainty \cite{Murphy2012MachineL}.

\begin{figure}[h]
\vspace{-3mm}
	\centering
	\includegraphics[width=.95\linewidth]{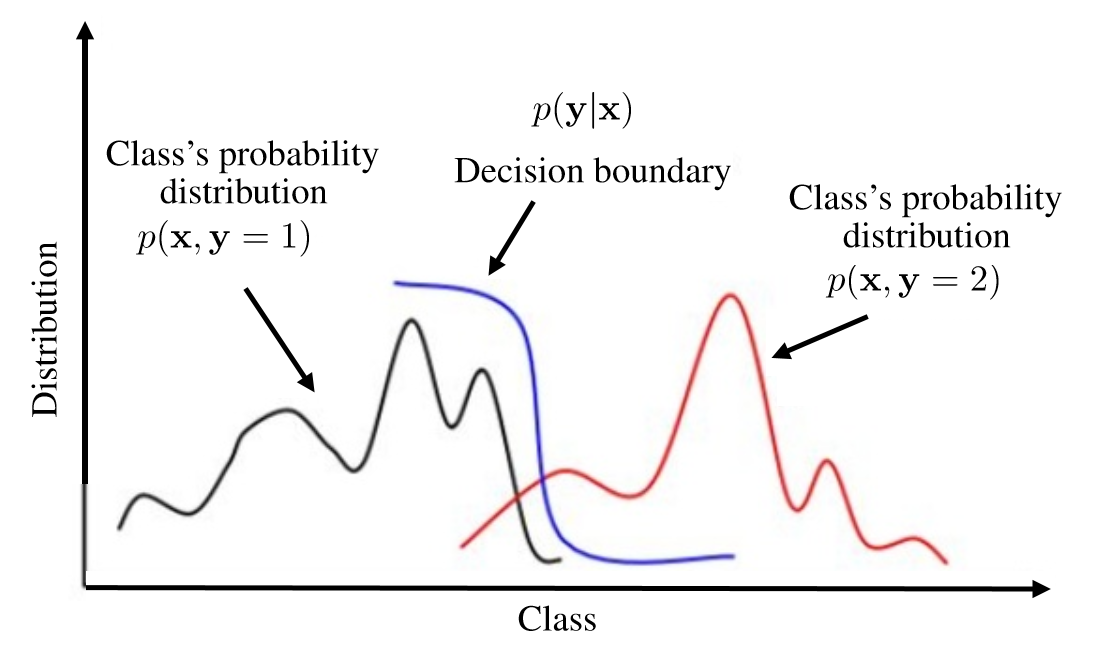}
	\vspace{-6mm}
	\caption{Differences between a generative and discriminative approaches; $\mathbf{x}$ is the input's features, and $y$ is the corresponding label}
	\label{fig:discGen}
	\vspace{-3mm}
\end{figure}

A learning method is \emph{generative} if it generates new data instances by approximating or learning the probability distribution that governs training set. In detail, it learns a joint probability distribution over the input's features. The \emph{naive Bayes} classifier is an example of generative models. A \emph{discriminative} model directly determines the label of an input instance by estimating a conditional probability for the labels given the input's features. \emph{Logistic regression} is an example of discriminative models. Discriminative models can only be used in supervised learning tasks, whereas generative models are employed in both supervised and unsupervised settings \cite{Ng2001OnDV}. Figure \ref{fig:discGen}, sketches the differences between the mentioned approaches; A discriminative model learns a decision boundary that separates the classes whereas a generative model learns the distribution that governs input data in each class.


\emph{Deep Neural Networks (DNNs)} are extremely powerful machine learning models that achieve excellent performance on difficult tasks such as speech recognition, machine translation, and visual object recognition \cite{Krizhevsky2012ImageNetCW, LeCun1998GradientbasedLA, Hinton2012DeepNN}. 
DNNs aim at learning feature hierarchies at multiple levels of abstraction that allow a system to learn complex functions mapping the input to the output directly from data, without depending completely on human-crafted features. Another advantage of DNNs over classical machine learning models is the notion of \emph{distributed representation} that gives rise to much better generalization on unseen instances \cite{HintonD.R.86}. The learning process in a DNN equals to estimating its parameters, that can be done it via \emph{Stochastic Gradient Descent (SGD)} \cite{Goodfellow-et-al-2016}.

\emph{Recurrent Neural Networks (RNNs)} are a family of DNNs with \emph{cyclic} structures that make them suitable for processing sequential data of variable length.
 RNNs suffer from \emph{catasrophic forgetting}, i.e., the model forgets the learned patterns, and \emph{optimization instability}, i.e., the optimization does not converge \cite{Goodfellow-et-al-2016}. To alleviate the former one can use the \emph{Long Short-Term Memory (LSTM)} architecture \cite{LSTM1997}, and to fix the latter \emph{gradient clipping} \cite{Pascanu2012OnTD} can be used.

\begin{figure}[h]
\vspace{-3mm}
	\centering
	\includegraphics[width=1\linewidth]{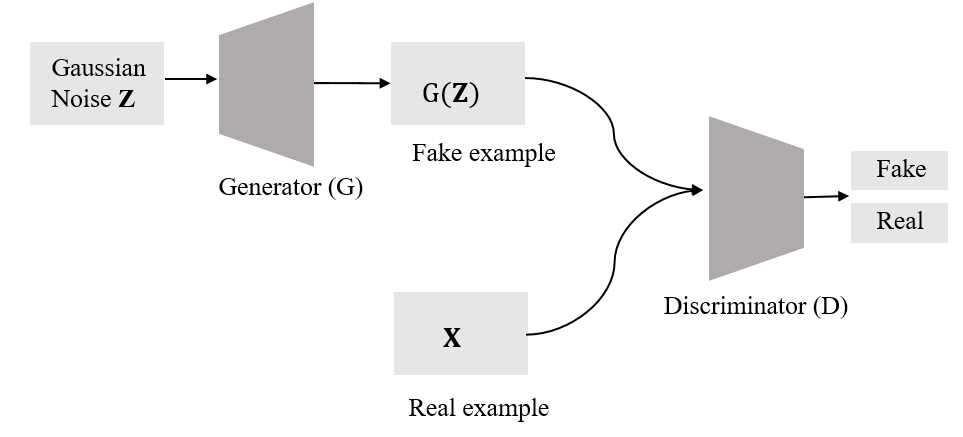}
	\caption{Generative adversarial nets \cite{GANNIPS2014_5423}; the generator produces fake examples from Gaussian noise, and the discriminator determines which of its input is real or fake.}
	\label{fig:vanila gan}
	\vspace{-3mm}
\end{figure}

\noindent \emph{Generative Adversarial Nets (GANs)} \cite{GANNIPS2014_5423} are an example of generative models. It takes a training set, consisting of samples drawn
from a distribution $p_{data}$, and learns to represent an estimate of that distribution, which results in distribution $p_{model}$. GANs employ two neural network models, called players, simultaneously, see Fig. \ref{fig:vanila gan}.
The two players correspond to a \emph{generator} and a \emph{discriminator}. The generator takes samples from a simple distribution, e.g., \emph{Gaussian noise}, to produce instances, i.e., \emph{fake instances}, which are similar to input instances, i.e., \emph{real instances}. The discriminator is a binary classifier such as logistic regression whose job is to distinguish real instances from generated instances, i.e., fake instances. The generator tries to create instances that are as realistic as possible; its job is to fool the discriminator, whereas the discriminator's job is to identify the fake instances irrespective of how well the generator tries to fool it. It is an adversarial game because each player wants to maximize its own outcome which results in minimization of the other player's outcome.  The game finishes when the players reach to \emph{Nash equilibrium} that determines the optimal solution. In the equilibrium point the discriminator is unable to distinguish between real and fake instances.

\subsection{Predictive Process Monitoring of Suffix Generation and Remaining Time}
\label{subsec: related work}
\noindent This section reviews works on predictions for an ongoing case, i.e., next event, suffix, and remaining time, using deep learning techniques.

The work by Evermann et al. \cite{EVERMANN2017129} uses the LSTM architecture for the next activity prediction of an ongoing trace. It uses embedding techniques to represent categorical variables. The authors used a technique called \emph{process hallucination} to generate the suffix and the remaining time  for an ongoing case, i.e., a process model execution.



Tax et al. \cite{Tax17} propose a similar architecture based on LSTMs. This work uses a one-hot vector encoding to represent categorical variables. Suffix prediction is made by next activity predictions iteratively using \emph{arg-max} selection method. The proposed approach outperforms \cite{EVERMANN2017129}.

Camargo et al. \cite{Camargo2019LearningAL} employ a composition of LSTMs and feedforward layers to predict the next activity and its timestamp and the remaining cycle time and suffix for a running case. The approach uses embedding techniques similar to \cite{EVERMANN2017129}. Similar to \cite{EVERMANN2017129} it uses process hallucination to generate the suffix and the remaining time  for an ongoing case. 


Lin et al. \cite{Lin2019MMPredAD} propose an sequence to sequence framework based on LSTMs to predict the next activity and the suffix of an ongoing case where It uses all available information in input log, i.e., both control-flow and performance attributes. The experimental setup of this approach is different from \cite{EVERMANN2017129, Tax17, Camargo2019LearningAL}. 


Taymouri et al. \cite{Taymouri2020PredictiveBP} propose a GAN architecture for the next activity and timestamp prediction. The proposed architecture invokes an LSTM for both the discriminator and the generator. 
It uses one-hot vector encoding to deal with categorical variables. The results showed that this technique outperforms previous techniques \cite{Tax17,EVERMANN2017129, Camargo2019LearningAL, Lin2019MMPredAD, Pasquadibisceglie2019UsingCN} for the next activity and timestamp prediction.

\begin{figure*}[h]
\vspace{-5mm}
	\centering
	\includegraphics[width=.8\linewidth]{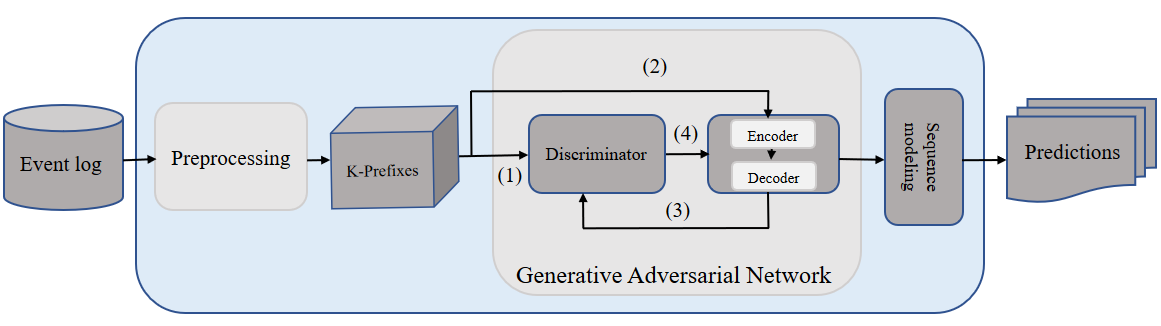}
	\vspace{-3mm}
	\caption{Overall approach for suffix and remaining time prediction}
	\label{fig:framework}
	\vspace{-3mm}
\end{figure*}

\section{Approach}
\label{sec:proposed approach}

\noindent The main aim of predictive process monitoring is to predict the corresponding attributes of ongoing process executions one or a few steps ahead of time. This paper, for an ongoing process execution (prefix), predicts the suffix, and remaining cycle time, i.e., the sequence of next activity labels and their timestamps until the case finishes, respectively.

To achieve this goal, we propose a framework that exploits both traditional training,i.e., gradient-based, and adversarial training inspired by GANs \cite{GANNIPS2014_5423, arjovsky2017wasserstein} in a novel way for the process mining context, see Fig. \ref{fig:framework}. It has three parts as follows:

\emph{Data prepossessing}: It prepares the input data in the form of prefixes and suffixes for the prediction task. It uses one-hot encoding to represent categorical variables. \emph{Adversarial predictive process monitoring net} is made of two neural networks, i.e., generator, and discriminator, where the former generates suffixes and the latter discriminates the ground truth and predicted suffixes.
The generator itself has two internal parts called \emph{encoder} and \emph{decoder}. The encoder maps a prefix into a fixed-size vector upon which the decoder generates a suffix. The minmax game between generator and discriminator starts
by proposing fake and real suffixes. Real suffixes are those in the training set, and fake suffixes are formed from the generator's output.
The training runs as a game between two players, where the generator's goal is to maximize
the quality of generated suffixes to fool the discriminator, and the discriminator's goal is to minimize its error in distinguishing real and fake suffixes, see flows (1), (2) in Fig. \ref{fig:framework}. It is an adversarial game since the generator and the discriminator compete with each other, i.e., learning from the opponent's mistake, see flows (1), (3) in Fig. \ref{fig:framework}. Thus maximizing one objective function minimizes the other one and vice versa. Note that,
the discriminator can send feedback to the generator if suffixes are differentiable; however in our work a suffix is a sequence of categorical items. Thus, we get a continuous approximation to each suffix using Gumbel-Softmax distribution. After training, the \emph{Sequence modeling} part adopts a best-first search, a.k.a., \emph{beam search}, over the generator's output such that for a prefix, it provides the $k$ most probable suffixes.

The contributions of this paper compared to the existing works are as follows:

\begin{itemize}
        \item The works in \cite{Tax17, Lin2019MMPredAD, Camargo2019LearningAL} train a neural network for the next activity and timestamp predictions. Thus, for an ongoing case, a suffix is generated by predicting the next event iteratively until the case finishes. This approach propagates the error from one step to the future steps, which results in poor suffixes. In this paper, we propose an end-to-end approach where a model is trained on each pair of prefix and suffix in training set directly via a minmax game. Thus, the model learns the existing relationships among activities in prefixes and suffixes.  
        
    \item The generator of GAN architecture in Taymouri et al. \cite{Taymouri2020PredictiveBP}, and the frameworks in \cite{Tax17, Lin2019MMPredAD, Camargo2019LearningAL} are RNNs (LSTM architecture). RNNs cannot map an input sequence into an output sequence of different lengths. In our work, the generator is made of two separate RNNs (LSTM architecture), i.e., encoder and decoder, where it allows mapping prefixes of variable lengths into suffixes of variable lengths.
    
    
    \item GANs work with differentiable samples such as images. However, in our framework, a suffix (sample) is a sequence of categorical items which
    is not differentiable. Thus, it doesn't allow the generator to receive feedback from the discriminator's output. We get a differentiable suffix, from the \emph{Gumbel-Softmax} distribution \cite{jang2016categoricalGumble, goodfellow2016nips} to make a continuous approximation to a sequence of categorical items.
    
    
    \item In many real-world applications, having a set of ranked suffixes for a prefix is beneficial. The works in \cite{Tax17, Camargo2019LearningAL, EVERMANN2017129,Lin2019MMPredAD} generate only one suffix for an input prefix. 
    In contrast, in this work, we apply a best-first search, on top of the generator's output to get the $k$ most probable suffixes. These suffixes, later, can be used for further analysis and plannings.
 
\end{itemize}

We do not limit the above contributions to the predictive process monitoring tasks, and one can adopt specialized GAN-like frameworks to the full range of process mining applications. We organize the rest of this section as follows. First, the preliminary definitions are presented. Following that, we formalize the required data prepossessing. Next, an in-depth explanation of Gumbel-Softmax reparameterization trick will be presented. After that, RNNs and the encoder-decoder architecture used in our framework will be provided. Finally, we give details of the adversarial predictive process monitoring net, including its training and optimization.

\subsection{Preliminaries and Definitions}
\label{sec: preliminaries}
\noindent This section provides the required preliminaries and definitions for the formalization of the proposed approach.



\begin{definition}[Gradient]
\label{def:grad}
For a function $f(\mathbf{x})$ with $f: \mathbb{R}^n \rightarrow \mathbb{R}$, the partial derivative $\frac{\partial}{\partial x_i} f(\mathbf{x})$ shows how $f$ changes as only variable $x_i$ increases at point $\mathbf{x}$. With that said, a vector containing all partial derivatives is called $gradient$, i.e., $\nabla_{\mathbf{x}} f(\mathbf{x}) = [ \frac{\partial}{\partial x_1} f(\mathbf{x}), \frac{\partial}{\partial x_2} f(\mathbf{x}),\dots,\frac{\partial}{\partial x_n} f(\mathbf{x})]^T$.
\end{definition}

\begin{definition}[Event, Trace, Event Log]
\label{def:trace event log}
An $event$ is a tuple $(a, c, t, (d_1, v_1), \ldots, (d_m, v_m))$ where $a$ is the activity name (label), $c$ is the case id, $t$ is the timestamp, and $(d_1, v_1) \ldots, (d_m, v_m)$ (where $m \geq 0$) are the event attributes (properties) and their associated values.
A $trace$ is a non-empty sequence $\sigma = \langle e_1,\ldots, e_{n} \rangle$ of events such that $\forall i,j \in \{1, \dots, n\} \; e_i{.}c = e_j{.}c$. An event log $L$ is a multiset  \{$\sigma_1, \ldots \sigma_n$\} of traces.
\end{definition}

\noindent A trace (process execution) also can be shown by a sequence of vectors, where a vector contains all or part of the information relating to an event, e.g., event's label and timestamp. Formally, $\sigma = \langle \mathbf{x}^{(1)},\mathbf{x}^{(2)}, \dots, \mathbf{x}^{(t)} \rangle$, where $\mathbf{x}^{(i)} \in \mathbb{R}^n$ is a vector, and the superscript shows the time-order upon which the events happened. 


\begin{definition}[k-Prefix (from beginning)]
\label{def: prefix}
Given a trace  $\sigma = \langle e_1,\ldots,e_{n} \rangle$, a $k$-$prefix$, $\sigma_{\leq k}$, is a non-empty sequence $\langle e_1,e_{2},\dots, e_{k} \rangle$.
\end{definition}

\noindent For $\sigma_{\leq k}$, the corresponding suffix is shown by $\sigma_{> k}$.
The above definition holds when an input trace is shown by a sequence of vectors. For example, given $\sigma = \langle \mathbf{x}^{(1)},\mathbf{x}^{(2)},\mathbf{x}^{(3)},\mathbf{x}^{(4)} \rangle$, $\sigma_{\leq 2} = \langle \mathbf{x}^{(1)},\mathbf{x}^{(2)} \rangle$, $\sigma_{> 2} = \langle \mathbf{x}^{(3)},\mathbf{x}^{(4)} \rangle$, and $\sigma_{\leq 3} = \langle \mathbf{x}^{(1)},\mathbf{x}^{(2)},\mathbf{x}^{(3)} \rangle$, $\sigma_{> 3} = \langle \mathbf{x}^{(4)} \rangle$.

\subsection{Data Preprocessing}
\label{subsec: data preprocessing}

\noindent 

The approach in this paper learns a function, $f$, that given a $k$-prefix, $\sigma_{\leq k}$, generates $f(\sigma_{\leq k}) = \hat{\sigma}_{>k}$ which is the prediction for the ground truth suffix $\sigma_{>k}$. $\hat{\sigma}_{>k}$ can be viewed as a sequence of next attributes until it reaches the end of the case. For the sake of simplicity, we only predict a sequence of event's label, i.e., \emph{activity}, and its \emph{timestamp}, see Def. \ref{def:trace event log}. An activity's timestamp is calculated as the time elapsed between the timestamp of one event and the event's timestamp that happened one step before. The \emph{remaining cycle time} of a prefix, is the amount of time that its suffix needs to finish, i.e., the sum of time elapsed for activities in $\sigma_{>k}$.



There are several methods in literature to encode and represent categorical variables. This paper, uses \emph{one-hot encoding} due to its simplicity, and to manifest the viability of the proposed architecture does not owe to the data representation part. Indeed, one can integrate various embedding representations.


The one-hot vector encoding of a categorical variable is a way to create a binary vector (except a single dimension which is one, the rest are zeros) for each value that it takes. Besides, we use $\langle \mathrm{EOS} \rangle $ to denote the end of a trace. 
\noindent For example, lets $\mathcal{E} = \{a_1, a_2, a_3, a_4, a_5, \langle \mathrm{EOS} \rangle\}$ shows the set of activities name including $\langle \mathrm{EOS} \rangle $, and $\sigma = \langle a_1 ,a_3, a_4, \langle \mathrm{EOS} \rangle \rangle$. The one-hot vector encoding of $\sigma$ is the following sequence of vectors:

\vspace{-3mm}
\footnotesize{\begin{equation*}
 \langle (\underbrace{1,0,0,0,0,0}_\text{$a_1$}), (\underbrace{0,0,1,0,0,0}_\text{$a_3$}), (\underbrace{0,0,0,1,0,0}_\text{$a_4$}), (\underbrace{0,0,0,0,0,1}_\text{$\langle EOS \rangle$})   \rangle  
\end{equation*}}
\normalsize
\noindent Furthermore, if $\mathbf{x}^{(i)}$ shows the one-hot vector of  $e_i.a$, then, one can \emph{augment} the former with the other attributes of the latter. In this paper, we augment one-hot vectors with the time elapsed between the timestamp of one event and the event’s timestamp time that happened one step before.


For an event log containing $n$ traces, we split it into  80:15:5 percentages for train, test, and validation sets respectively. Formally, an instance in each of training, test, and validation set is a pair of prefix and suffix like $(\sigma_{\leq k}, \sigma_{>k})$ for $2 \leq k \leq |\sigma|-1$, and $\sigma_{\leq k}$ and $\sigma_{>k}$ are sequences of augmented one-hot vectors.

\subsection{Categorical Reparameterization with Gumbel-Softmax}
\label{subsec: gumble softmax}
This section shows how to approximate a differentiable sample from a multinomial distribution which will be used in our framework. 

Lets assume that a categorical variable $x$ has $k$ different values, i.e., classes or activity names, with probabilities $\boldsymbol{\pi} = [\pi_1,\pi_2,\dots,\pi_k]^T$. We assume
each categorical sample is shown by a $k$-dimensional one-hot vector. The probabilities $\pi_i$ could be, for example, the outputs of a neural network for the prediction task.
A simple way to draw a categorical sample $\mathbf{z}$ from class probabilities $\pi_1,\pi_2,\dots,\pi_k$ is to apply \emph{Gumbel-Max} trick \cite{gumbel1954statistical,NIPS2014_5449Gumble} as follow:
\begin{equation}
\label{eq:gumble max}
    \small{ \mathbf{z} = one\text{-}hot\left( arg \operatorname*{max}_{i}[g_i + log( \pi_i)]    \right) }
\end{equation}

\noindent Where $g_i$ are i.i.d samples from Gumbel(0,1). The calculated $\mathbf{z}$ could be, for example, the output of a neural network. 

Note that, the generated samples according to Eq. \ref{eq:gumble max} are not differentiable because \emph{arg max} operator outputs a discrete value. In detail, the gradient of $\mathbf{z}$ with respect to each $\pi_i$ is  zero, i.e., $\nabla_{\boldsymbol{\pi}} \boldsymbol{z} =[  \frac{\partial}{\partial \pi_1} \boldsymbol{z},\dots, \frac{\partial}{\partial \pi_k} \boldsymbol{z}]^T=[0,\dots,0]^T$. 

One can get a continuous, differentiable approximation to \emph{arg max} operator by employing \emph{Softmax} operator to generate samples $\mathbf{y}$ whose elements, i.e., $y_i$, are computed as follows:

\begin{equation}
\label{eq: gumble softmax}
   \small{ y_i = \frac{exp((log (\pi_i) + g_i)/\tau)}{\sum_{j=1}^k exp((log (\pi_j) + g_j)/\tau)} }
\end{equation}

\noindent Where $\tau$ is a parameter called \emph{temperature}. When $\tau \rightarrow 0$, the samples generated by Eq. \ref{eq: gumble softmax} have the same distribution as those generated by Eq. \ref{eq:gumble max}, and when $\tau \rightarrow \infty$ the samples are always the uniform probability vector.

\begin{figure}[h]
\vspace{-3mm}
	\centering
	\includegraphics[width=.8\linewidth]{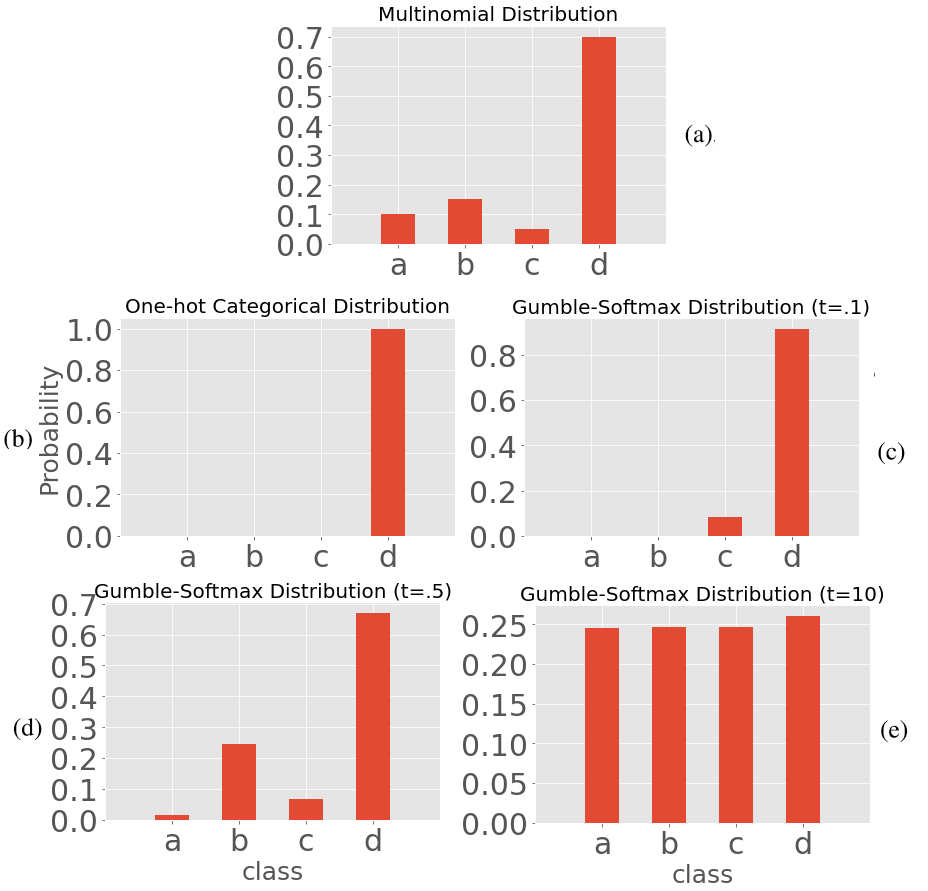}
	\vspace{-3mm}
	\caption{Gumbel-Softmax distribution sample with different $\tau$ values versus one-hot categorical sample}
	\label{fig: gumbel soft example}
	\vspace{-4mm}
\end{figure}

For example, assume that there is a probability distribution $\boldsymbol{\pi}= [0.1, 0.15,0.05,0.70]^T$ over classes \emph{a,b,c,d}, as shown in Fig. \ref{fig: gumbel soft example} (a). If one samples according to Eq. \ref{eq:gumble max}, then \emph{arg max} function selects \emph{d} which results in a one-hot vector $\mathbf{z} = [0,0,0,1]^T$, see Fig. \ref{fig: gumbel soft example} (b). In contrast, if one uses Eq. \ref{eq: gumble softmax} with $\tau =0.1, 0.5$ and $10$, the generated vectors are $[\num{6.9e-7}, \num{4.9e-6}, \num{8.3e-2}, 0.91]^T$, $[0.015, 0.247, 0.066, 0.670]^T$, and $[0.245, 0.246, 0.246, .260]^T$ respectively, see Fig. \ref{fig: gumbel soft example} (c), (d), and (e).

\subsection{Recurrent Neural Networks and Encoder-Decoder Architecture}
\label{subsec: RNN graphical model}
\noindent This section outlines RNNs, although, the concepts hold for any RNN architectures such as LSTM. Next, we provide the encoder-decoder architecture in details.




\begin{figure}[h]
\vspace{-1mm}
	\centering
	\includegraphics[width=1\linewidth]{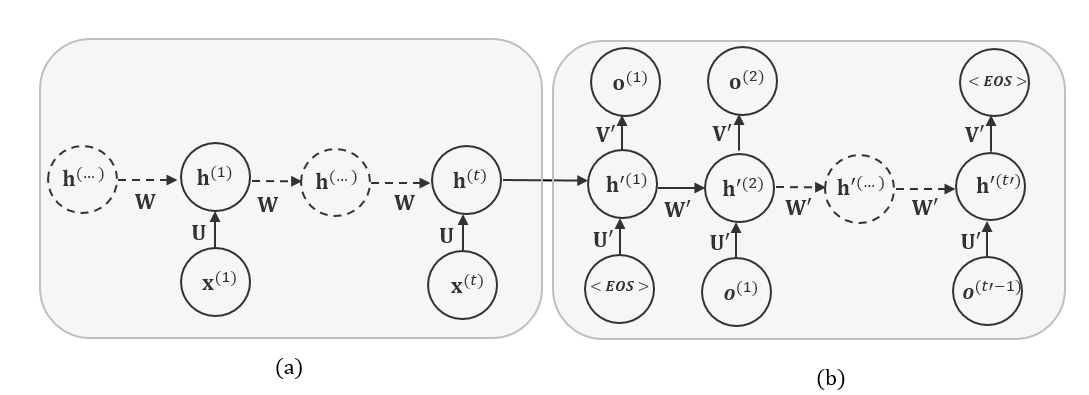}
	\vspace{-4mm}
	\caption{Encoder-Decoder architecture; For the sake of simplicity we hide vectors $\mathbf{c}$,$\mathbf{b}$, and functions $\phi_h$, $\phi_o$ (a) Encoder maps an input prefix into a fixed-size vector $\mathbf{h}^{(k)}$, (b) Decoder receives $\mathbf{h}^{(k)}$ and $<EOS>$ by which it generates s suffix}
	\label{fig:encoderDecoeer}
	\vspace{-3mm}
\end{figure}

 Given a sequence of inputs $\langle \mathbf{x}^{(1)},\mathbf{x}^{(2)}, \dots, \mathbf{x}^{(k)} \rangle$, an RNN computes sequence of outputs $\langle \mathbf{o}^{(1)},\mathbf{o}^{(2)}, \dots, \mathbf{o}^{(k)} \rangle$ 
via the following recurrent equations:

\footnotesize{\begin{flalign}
\begin{split}
\label{eq:rnn}
    \mathbf{h}^{(t)} = \phi_h(\mathbf{W}^T\mathbf{h}^{(t-1)} + \mathbf{U}^T \mathbf{x}^{(t)}+\mathbf{c}) ,  \\
    \mathbf{o}^{(t)} = \phi_o(\mathbf{V}^T\mathbf{h}^{(t)} + \mathbf{b}),\quad \forall t \in \{1,2,\dots,k\} 
\end{split}
\end{flalign}}
\normalsize
\vspace{-2mm}

\noindent Where $\mathbf{o}^{(t)}$ is the RNN's output for ground truth vector $\mathbf{y}^{(t)}$; $\phi_h$ and $\phi_o$ are nonlinear element-wise functions, and the set $\theta = \{\mathbf{W}, \mathbf{U}, \mathbf{V}, \mathbf{c, b}\}$, is the network's parameters.


According to Eq. \ref{eq:rnn}, one sees that both input and output sequences have the same length, i.e., $k$, which poses a limitation for generating a suffix for an input prefix, as they have different lengths. A simple strategy for general sequence learning is to map the input sequence, i.e., prefix, to a fixed-sized vector using one RNN, called encoder, and then map it to the target sequence, i.e., suffix, with another RNN, called decoder. The encoder-decoder architecture \cite{Sutskever2018321V, cho2014learningEncoderDecoder}, as depicted in Fig. \ref{fig:encoderDecoeer}, allows the input and output sequences to have variable lengths. 

Formally, suppose that the input prefix and the output suffix are $\sigma_{\leq k} = \langle \mathbf{x}^{(1)},\mathbf{x}^{(2)}, \dots, \mathbf{x}^{(k)} \rangle$, and $\sigma_{> k} = \langle \mathbf{y}^{(1)},\mathbf{y}^{(2)}, \dots, \mathbf{y}^{(k')} \rangle$, respectively. Note that the prefix and the suffix have different lengths. The goal of the encoder-decoder architecture is to learn a probability distribution $p_m$ to maximize $p_m(\mathbf{y}^{(1)},\mathbf{y}^{(2)}, \dots, \mathbf{y}^{(k')} | \mathbf{x}^{(1)},\mathbf{x}^{(2)}, \dots, \mathbf{x}^{(k)})$. To achieve this goal, the encoder creates a fixed-size vector $\mathbf{h}^{(k)}$, which is its last hidden state, from $\sigma_{\leq k}$ using Eq. \ref{eq:rnn}. Then, the decoder computes the mentioned conditional probability as follows:

\vspace{-3mm}
\footnotesize{\begin{flalign}
 \begin{split}
\label{eq: decoder probability}
    p_m(\mathbf{y}^{(1)},\mathbf{y}^{(2)}, \dots, \mathbf{y}^{(k')} | \mathbf{x}^{(1)},\mathbf{x}^{(2)}, \dots, \mathbf{x}^{(k)}) = \\ \displaystyle\prod_{t=1}^{k'} p_m(\mathbf{y}^{(t)}|\mathbf{h}^{(k)},\mathbf{y}^{(1)},\dots, \mathbf{y}^{t-1)})
\end{split} 
\end{flalign}}
\normalsize
\vspace{-2mm}

\noindent Each $p_m(\mathbf{y}^{(t)}|\mathbf{h}^{(k)},\mathbf{y}^{(1)},\dots, \mathbf{y}^{t-1)})$ distribution is represented with a Softmax function over $\mathbf{o}^{(t)}$. The suffix generation by the decoder continuous until it reaches $<EOS>$. Finally, the end-to-end training is to minimizing the negative log probability of Eq. \ref{eq: decoder probability} for every pair of prefix and suffix in the training set $\mathcal{S}$:
\begin{equation}
   - \displaystyle\sum_{(\sigma_{\leq k}, \sigma_{> k}) \in \mathcal{S}} log (p_m(\sigma_{> k}|\sigma_{\leq k}))
    \label{eq: encoder decoder eq}
\end{equation}

After training, decoder generates a sequence of outputs $\langle \mathbf{o}^{(1)},\mathbf{o}^{(2)}, \dots, \mathbf{o}^{(k')} \rangle$ for a prefix. We get a suffix, i.e., $\hat{\sigma}_{>k}$, by applying \emph{one-hot} and \emph{arg max} function on each $\mathbf{o}^{(t)}$ for $t \in \{1,\dots,k'\}$ properly. Note that, the final output, i.e., the suffix, is a sequence of augmented one-hot vectors.



\begin{figure}[h]
\vspace{-2mm}
	\centering
	\includegraphics[width=1\linewidth]{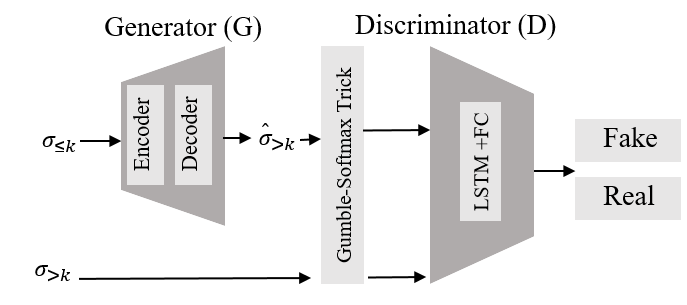}
	\vspace{-7mm}
	\caption{Proposed GAN architecture for suffix generation}
	\label{fig:gan}
	\vspace{-4mm}
\end{figure}

\subsection{Adversarial Predictive Process Monitoring Nets}
\label{subsec: proposedGAn}
\noindent This section presents the core contribution of this paper by proposing an adversarial process to estimate a generative model for the predictive process monitoring tasks. 



In the proposed adversarial architecture, shown in Fig. \ref{fig:gan}, the generator, denoted by $G(;\mathbf{\theta}_g)$, is an encoder-decoder architecture as defined in Sec. \ref{subsec: RNN graphical model}, and the discriminator, denoted by $D(;\mathbf{\theta}_d)$, is composed of LSTM followed by a fully connected layer. In detail, given a prefix, i.e., $\sigma_{\leq k}$, the generator's output is a suffix, i.e., $G(\sigma_{\leq k})= \hat{\sigma}_{> k}$. The Gumbel-Softmax part of Fig. \ref{fig:gan}, creates  continuous differentiable approximations for the one-hot encoding parts of ground truth and predicted suffixes $\sigma_{> k}$ and $\hat{\sigma}_{> k}$ as explained in Sec. \ref{subsec: gumble softmax}. Finally, the discriminator receives the outputs of the Gumbel-Softmax part and decides whether the suffixes are the same by assigning a probability to them, i.e., $0$ for $\hat{\sigma}_{> k}$, and $1$ for $\sigma_{> k}$.

In the proposed architecture, $\sigma_{> k}$, and $\hat{\sigma}_{> k}$ are considered as real and fake suffixes, respectively. We want the $G$'s output, i.e., $\hat{\sigma}_{> k}$, to be as close as possible to ground truth $\sigma_{> k}$, such that, $D$ gets confused in discriminating the mentioned suffixes. Formally, for a pair $(\sigma_{\leq k}, \sigma_{>k}))$ according to the recent works in GANs \cite{Sonderby2016aGAN} the optimization game is as follows:
\begin{flalign}
\begin{split}
\label{eq: gan equation}
    \mathcal{L}(D;G) = -log(D(\sigma_{>k})) - log(1-D(G(\sigma_{\leq k}))) \\
    \mathcal{L}(G;D) = -\left [ log(D(G(\sigma_{\leq k})))  - log(1-D(G(\sigma_{\leq k}))) \right ]
\end{split}
\end{flalign}

\noindent Where $\mathcal{L()}$ is the loss value. Equation \ref{eq: gan equation} originally  was proposed for synthesising images, however, we apply it for generating suffixes. It iterates two steps:  first, it updates discriminator $D$ by lowering  $\mathcal{L}(D;G)$, keeping $G$ fixed, then it updates $G$ by lowering $\mathcal{L}(G;D)$ keeping $D$ fixed. It can be shown that the optimization in Eq. \ref{eq: gan equation} amounts to minimizing the distance between two probability distributions that generate the ground truth suffixes and fake suffixes, respectively \cite{Sonderby2016aGAN}. 


\begin{algorithm}
\small{\SetAlgoLined
 initialization parameters of $D$ and $G$ with standard normal distribution\;
 \While{Number of iterations}{
 \For{ each $(\sigma_{\leq k}, \sigma_{>k}) \in \mathcal{S}$}{
 update $\theta_d$ by minimizing $L(D;G)$\;
 update $\theta_g$, by minimizing $L(G;D)$ + $L_{supervised}$\;
 }
 } 
 }
 \caption{Proposed adversarial training, MLMME}
 \label{alg: ganTraining}
\end{algorithm}

\begin{algorithm}
\small{\SetAlgoLined
 initialization parameters of $D$ and $G$ with standard normal distribution\;
 \While{no convergence}{
 \For{ each $(\sigma_{\leq k}, \sigma_{>k}) \in \mathcal{S}$}{
 update $D$'s parameters, $\theta_d$, by minimizing $\mathcal{L}(D;G)$\;
 update $G$'s parameters, $\theta_g$, by minimizing $\mathcal{L}(G;D)$ + Eq. \ref{eq: encoder decoder eq}\;
 }
 } 
 }
 \caption{Training proposed adversarial net}
 \label{alg: ganTraining}
\end{algorithm}

The training of the proposed framework is shown in Alg. \ref{alg: ganTraining}. For each pair of prefix and suffix $(\sigma_{\leq k}, \sigma_{>k})$ in the training set $\mathcal{S}$, we update the parameters of the discriminator and the generator according to Eq. \ref{eq: gan equation}. Also, the parameters of the latter are further updated using Eq. \ref{eq: encoder decoder eq}, to speed up the training convergence. Training GANs is a difficult task because of the optimization game, i.e., updates of one player can completely undo the other player's updates; thus, no convergence is guaranteed, or it needs many iterations \cite{Goodfellow2017NIPS2T}.
Theoretically, convergence happens when the discriminator is unable to distinguish the real and fake suffixes. Formally, it takes place when $E(\mathcal{L}(G;D)) \rightarrow 0$, or equivalently, $D(G(\sigma_{\leq k})) \rightarrow 0.5$,  for training data. Note that, one can use Alg. \ref{alg: ganTraining} with mini-batch to speed up training time.

After training, we disconnect the discriminator and use the generator for the suffix generation task.

\subsection{Sequence Modeling}
\label{subsec: sequence modeling}
This section explains a  breadth-first search, a.k.a., beam search \cite{BeamSEARCHLowerre1980TheHS}, for generating a suffix, given probability distribution over the activity names. The probability distribution could be, for example, the neural network's output.
Let's assume that there is a sequence of probability distributions $[0.3,0.35,0.3, 0.05]^T, [0.35,0.3,0.3, 0.05]^T$, and $[0.05,0.3,0.35, 0.3]^T$ for three timestamps over four activities \emph{a,b,c}, and \emph{d}, respectively. A suffix can be made as follows:
\begin{itemize}
    \item \emph{Arg max selection}: This approach selects an element with maximum probability in each timestamp, resulting in $bac$ \cite{Tax17, Lin2019MMPredAD}. 
    \item \emph{Random selection}: This approach, a.k.a., process hallucination, selects an activity randomly according to the probability distribution in that timestamp. Different runs results in various suffixes e.g., $acd,aac,bba,$etc, \cite{Camargo2019LearningAL, EVERMANN2017129}.
\end{itemize}

Given a prefix, the first method is a greedy way to generate the most probable suffix given the trained model. However, it might not be the enough solution in many real-world applications where having different predictions is beneficial. 
The second method is more flexible, where, for a prefix, one can run the model several times and get several suffixes. It generates suffixes randomly according to some distribution; however, it does not control the quality of the generated suffixes, see the examples.
It becomes reliable, specifically for prediction tasks when the probability distribution is very sharp around one activity in each timestamp, which is not guaranteed in real-world applications. This approach needs very large number of simulations (runs) upon which an inference can be made \cite{SequenceMemozieNIPS2010_3938}.

In this work, we use a  breadth-first search, i.e., Beam search, that builds a search tree. The root is an empty state, i.e., suffix, and leaves are goals or solutions. A new state, i.e., suffix, can be made from the current state by concatenating an activity to the end of the current state's suffix. An intermediate state of a search tree is a suffix from the beginning, i.e., root, up to that state.  A state is a goal or solution if its suffix reaches $<EOS>$. At each level of the tree, we generate all the states' successors at the current level, sorting them in increasing order of a cost function, e.g., the sum of negative log probabilities, and select a predetermined number $k$, called the beam width, at each level. Only those $k$ states are expanded next. The time and space complexities of Beam search with width $k$ are $\mathcal{O}(kb)$ and $\mathcal{O}(k)$, respectively, where $b$ is the branching factor or the number of activities in our work.

A Beam search with width $k$, returns the $k$ most probable suffixes for input prefix. It is worth noting that \emph{Arg max} selection method is a Beam search with $k=1$, and process hallucination method is a Beam search with $k=\infty$ that selects a leaf randomly. For the above example a beam search with $k=1$ returns $bac$ with score $-(log(.35)+log(.35)+log(.35)) = 3.149$, and with $k=3$ the top 3 probable suffixes are $bac, bab$, and $bad$ with scores $3.149, 3.303$, and $3.303$, respectively.

\begin{table*}[h]
\vspace{-5mm}
	\centering 
	\footnotesize{ \begin{tabular}{|p{2.4cm}| p{1.2cm}| p{1.3cm} | p{1.0cm} | p{1.0cm} | p{1.2cm}| p{1.3cm}|p{1.0cm}|p{1.0cm}|}
		\hline
    \multicolumn{1}{|c|}{} & \multicolumn{4}{|c|} {\textbf{Average similarity $d$}} & \multicolumn{4}{|c|}{\textbf{ MAE (day)}}\\
		\hline \hline
		\textbf{Approach} & Helpdesk  & BPI12(W) & BPI12 & BPI17 & Helpdesk  & BPI12(W) & BPI12 & BPI17\\ [1ex]
		\hline
		Ours ($k=1$) & \textbf{0.8411}  &  \textbf{0.2662}    & \textbf{0.3326}  & \textbf{0.3361}  & \textbf{6.21}  & \textbf{12.12} & \textbf{13.62}  & \textbf{13.95} \\
		
		
		\hline
		\hline
		Taymouri et al. \cite{Taymouri2020PredictiveBP} &0.8089  &  0.3520  & 0.2266  & 0.2958\footnotemark[3]  & 6.30  & 34.56 & 169.23  &  80.81 \\
		Tax et al. \cite{Tax17} &0.7670  &  0.0632  & 0.1652  & 0.3152  & 6.32  & 50.11 &  380.1  & 170 \\
		Lin et al. \cite{Lin2019MMPredAD}\footnotemark &0.8740  &  N/A  & 0.2810  & 0.3010  & N/A  & N/A &N/A  & N/A\\
		\hline
	\end{tabular} }
		\caption{Average similarity $d$ for suffix generation (the larger, the better), and MAE for the remaining cycle time. $k$ is the beam width.}
	\label{table:accuracy and MAE}
		\vspace{-\baselineskip}
\end{table*}
\addtocounter{footnote}{-1}
\footnotetext[2]{This approach for training and testing considers different settings. For \emph{Helpdesk}, it uses prefixes longer than 3; for the other logs, prefixes are equal or longer than 5. These settings impact positively on the results, making this approach not comparable with the others used in Table \ref{table:accuracy and MAE}.}
\footnotetext[3]{For this dataset the tool ran out of memory in middle of training. Testing is done via the last saved checkpoint.}

\section{Evaluation}
\label{sec: evaluation}

\noindent We implemented our approach in Python 3.6 via PyTorch 1.2.0 on a single NVIDIA P100 GPU with 24 GB of RAM and CUDA 10.1, on Google Cloud. 
We used this prototype tool to evaluate the approach over four real-life event logs, against three baselines \cite{ Tax17, Lin2019MMPredAD, Taymouri2020PredictiveBP}. The choice of the baselines was determined by experimental settings and the tool availability. 
For this reason, we excluded from the experiment the work by Camargo et al. \cite{Camargo2019LearningAL} and Evermann et al. \cite{EVERMANN2017129} due to their evaluation method, i.e., \textit{process hallucination}, which suffers from randomness, see Sec.\ref{subsec: sequence modeling}, leading to unreliable results. The other baselines use \emph{Arg max} as the selection method. 
It must be noted that the settings used by Lin at al. \cite{Lin2019MMPredAD} tend to advantage this approach over the others, as the training and testing are biased towards longer prefixes. In fact, short prefixes are excluded based on a minimum length threshold (the specific settings are discussed in a footnote of Table \ref{table:accuracy and MAE}).

\subsection{Experimental Setup}
\noindent \textbf{Datasets:} The experiments were conducted using four publicly-available real-life logs obtained from the 4TU Centre for Research Data.\footnote{https://data.4tu.nl/repository/collection:event\_logs\_real} 
Table \ref{table:dataset}
shows the details of these logs.

\begin{itemize}
\vspace{-2pt}
    \item \textbf{Helpdesk}: It contains traces from a ticketing management process of the help desk of an Italian software company. 
    \item \textbf{BPI12}: It contains traces of a loan application process at a  Dutch financial institute. This process includes three sub-processes from which one of them is denoted as $W$ and used already in \cite{Tax17}. As such, we extract it from this dataset: BPI12(W).
    \item \textbf {BPI17}: It contains traces of a loan application process at the same Dutch financial institute but in a different period of time.
\end{itemize}

\begin{table}[h]
    \centering
    \vspace{-2mm}
\footnotesize{\begin{tabular}{|p{1cm} |p{.6cm}|p{.9cm}|p{.8cm}|p{1.3cm}|p{1.5cm}|}
  \hline
  \textbf{Log}  & \textbf{Traces} & \textbf{Events} & \textbf{Activity} & \textbf{Max/Min/ Avg $|\sigma|$} & \textbf{Avg./Max Cycle Time (day)} \\ 
  \hline
  Helpdesk & 3,804     & 13,087 & 9  & 14/1/3.60 & 8.79/ 55.9 \\
  BPI12 & 13,087     & 262,200 & 23  & 96/3/12.57 & 8.6/ 91.4\\
  BPI12(W) & 9,658     & 72,413 & 6  & 74/1/7.49  & 11.4/ 91 \\
  BPI17 & 31,509     & 1,202,267 & 24  & 54/7/4.59 & 21.8/ 169.1\\
  \hline
\end{tabular} }
\caption{Descriptive statistics of the datasets ($|\sigma|$ is the trace length)}
\vspace{-6mm}
\label{table:dataset}
\end{table}

\noindent \textbf{Evaluation measures:} For consistency, we reuse the same evaluation measures adopted in the baselines \cite{Tax17}. Specifically, to measure the accuracy of the generated suffix $\hat{\sigma}_{>k}$ against the ground truth $\sigma_{>k}$ we use $d(\hat{\sigma}_{>k},\sigma_{>k}) = 1- (D.L.(\hat{\sigma}_{>k},\sigma_{>k})/Max(|\hat{\sigma}_{>k}|,|\sigma_{>k}|) $, as the similarity measure, where $D.L.$ is Damerau–Levenshtein distance.
For the remaining time, we report Mean Absolute Error (MAE), i.e.\ the average of the absolute value between predictions and ground truths.\\

\noindent \textbf{Training setting:} For both generator (encoder, decoder) and discriminator we use a five layer LSTM with hidden size of 200. In addition, the discriminator is equipped with a fully connected layer for the binary classification task. In detail:
\begin{itemize}
\vspace{-3pt}
    \item Since the proposed method is generative, the training finishes when both the discriminator and the generator reach a convergence, as explained in Sec. \ref{subsec: proposedGAn}. In our experiments, we halted training after 500 epochs despite getting better results on the validation set with more epochs. 
    \item For each log, a training instance is a pair of prefix and suffix, where the prefix length is equal to or greater than 2. Formally, the training set consists of pairs $(\sigma_{\leq k}, \sigma_{>k})$ for $2 \leq k \leq |\sigma|-1$, where, $\sigma_{\leq k}$ and $\sigma_{>k}$ are sequences of augmented one-hot vectors (see Sec. \ref{subsec: data preprocessing}). 
    \item We use \emph{RMSprop} as an optimization algorithm for the proposed framework with learning rate $\num{5e-5}$. To avoid gradient explosion, we clip the gradient norm of each layer to 1. 
    \item We exponentially anneal the temperature $\tau$ of the Gumbel-Softmax distribution from 0.9 to 0 in Eq. \ref{eq: gumble softmax} to stabilize the training.
\end{itemize}

\noindent For the baselines, we used the best parameter settings, as discussed in the respective papers. 



\subsection{Results}
\label{subsec:results}

\noindent\textbf{Suffix Generation:} The second to fifth column of Table \ref{table:accuracy and MAE} show the average similarity $d$ of our approach and of the baselines, for each log. The first three rows represent the quality of the generated suffixes with different beam width $k$, where $k=1$ equals to the \emph{Arg max} selection method. We can see that for $k=1$, our approach provides considerably more accurate suffixes compared to the baselines for each dataset. From this table, we can draw several observations. First, the proposed framework outperforms the baselines for small-sized event logs, e.g., \emph{Helpdesk}, and $BPI12(w)$, but one sees its true merits for logs with long traces and a large set of activity names, e.g., $BPI12$ and $BPI17$.
This superiority amounts to the generator's internal architecture, i.e., encoder-decoder, and the use of a min-max (adversarial) game for training.
For $BPI12$, our approach achieves a significantly higher accuracy than the baselines, incl. Lin et al. \cite{Lin2019MMPredAD}, which uses different settings and employs other resource attributes (see Table \ref{table:accuracy and MAE}'s footnote). Second, one can see the quality of the generated suffixes improves as $k$ increases. Thus, for a prefix, one can have a ranked list of predicted suffixes based on their likelihood, which can be used for further analysis and planning.

\noindent\textbf{Remaining Time Prediction}:
 The last four columns in Table \ref{table:accuracy and MAE} show the MAE of the remaining time in days, for each log and for each approach, except \cite{Lin2019MMPredAD} as it does not support this prediction type. We can see that already for $k=1$, our approach provides much better accuracy than the baselines. Similar to suffix prediction, the results highlight the merit of our approach for logs with long traces like $BPI12$ and $BPI17$. 
 The higher MAE obtained by \cite{Tax17} depends on the error propagation issue and catastrophic forgetting, which results in generating long suffixes of repetitive events.
 
 Unlike the suffix generation part, larger $k$ values do not necessarily improve the MAE values. This is because the cost function in our beam search (see Sec. \ref{subsec: sequence modeling}) is the log-likelihood of a sequence of activities regardless of their timestamps. However, one can define a cost function that considers other attributes.
 
\begin{figure}[h]
\vspace{-2mm}
	\centering
	\includegraphics[width=.7\linewidth]{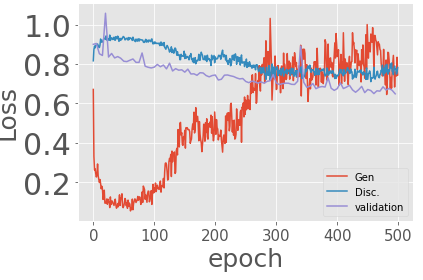}
	\vspace{-6mm}
	\caption{Proposed GAN architecture for suffix generation}
	\label{fig:convergence}
	\vspace{-4mm}
\end{figure}

\noindent\textbf{Behavior of the convergence}: We concluded our experiment by studying the convergence behavior of the generator and the discriminator while performing the minmax game in Alg. \ref{alg: ganTraining}. We observed one pattern in our experiments, as shown in Fig. \ref{fig:convergence}, which plots the generator loss on training/validation set, and the discriminator loss. One can observe that at the beginning the generator confuses the discriminator, and as the adversarial game progresses, the latter becomes stronger in subsequent epochs. However, this does not mean that the generator cannot generate accurate suffixes since the validation loss (purple line) decreases monotonically. Indeed, we observed that the generator produces partially correct suffixes, and improves itself over time. We stopped the training at 500 epochs, despite the downward trending in the generator's loss on the validation set. 

\section{Conclusion}
\label{sec: conclusion}
\noindent This paper put forward a novel generative adversarial framework for the suffix generation and the remaining time prediction, by adapting encoder-decoder Generative Adversarial Nets to the realm of sequential temporal data. The encoder-decoder architecture allows one to train the model directly on suffixes instead of training on the next event to generate suffixes. 
The training is achieved via a competition between two neural networks playing a minmax game. The generator maximizes its performance in generating accurate suffixes, while the discriminator minimizes its error in determining whether the generator's outputs are ground-truth sequences. At convergence, the generator confuses the discriminator in its task. We adopted Beam search on top of our architecture to provide the $k$ most probable suffixes for a prefix, which can be used for further analysis and planning.

The results of the experimental evaluation highlight the merits of our approach, which outperforms all the baselines, both in terms of similarity of the generated suffixes and the MAE of the prediction for the remaining time. 

Generative Adversarial Nets have received considerable attention both in academia and industry across many communities. 

\smallskip\noindent\textbf{Acknowledgments} This research is partly funded by the Australian Research Council (DP180102839).

\bibliographystyle{plain}
\bibliography{mybibfile}

\end{document}